# Improved phase-unwrapping method using geometric constraints

Guangliang Du[1], Minmin Wang[1], Canlin Zhou [1*],Shuchun Si[1], Hui Li[1], Zhenkun Lei[2],Yanjie Li[3]

[1] School of Physics, Shandong University, Jinan 250100, China

[2] Department of Engineering Mechanics, Dalian University of Technology, Dalian 116024, China

[3] School of Civil Engineering and Architecture, University of Jinan, Jinan, 250022, China

*Corresponding author: Tel: +8613256153609; E-mail address: canlinzhou@sdu.edu.cn

Conventional dual-frequency fringe projection algorithm often suffers from phase unwrapping failure when the frequency ratio between the high frequency and the low one is too large. Zhang et.al. proposed an enhanced two-frequency phase-shifting method to use geometric constraints of digital fringe projection(DFP) to reduce the noise impact due to the large frequency ratio. However, this method needs to calibrate the DFP system and calculate the minimum phase map $\varphi_{\min}$ at the nearest position from the camera perspective, these procedures are are relatively complex and more time-cosuming. In this paper, we proposed an improved method, which eliminates the system calibration and $\varphi_{\min}$ determination in Zhang's method,meanwhile does not need to use the low frequency fringe pattern. In the proposed method,we only need a set of high frequency fringe patterns to measure the object after the high frequency $\varphi_{\min}$ is directly estimated by the experiment. Thus the proposed method can simplify the procedure and improve the speed. Finally, the experimental evaluation is conducted to prove the validity of the proposed method.The results demonstrate that the proposed method can overcome the main disadvantages encountered by Zhang's method.

# 1. Introduction

Phase-based fringe projection is an important technique to perform three-dimensional (3D) shape measurements. It has been extensively investigated and widely used in many fields such as aerospace, biological engineering and military reconnaissance[1-4]. However, the phase obtained by the technique is wrapped, which means the value of the phase is contained within ($-\pi$, $\pi$). To unwrap the wrapped phase, many spatial and temporal phase unwrapping methods have been presented[5-7]. Spatial phase unwrapping is a process of integral accumulation, once the error appears, it will spread to the point around, so noise, shadow and discontinuous points in the actual measurement will affect the phase unwrapping quality[8-9]. Temporal phase unwrapping method can solve this problem, but it needs many frames of fringe images which would take much time [10]. To solve this problem, Liu[11] proposed the dual-frequency method, which combines a unit-frequency fringe pattern with a high-frequency. Liu[12] proposed tri-Frequency Heterodyne Method. Servin proposed a 2-step temporal phase unwrapping algorithm [13], which only needs the 2 extreme phase-maps to achieve exactly the same results as standard temporal unwrapping method. However, conventional dual-frequency phase unwrapping algorithm is greatly affected by the noise because the low-frequency phase must be with a unit frequency, which will lead to many errors when the frequency ratio is too big. In this regard, Zhang et al.[14] proposed FB-QGPUA. Different from the conventional dual-frequency method, the low-frequency fringe patterns do not have to be with a unit frequency, which makes the unwrapping process more robust because it decreases the frequency ratio,but the partially unwrapped phase need be unwrapped with quality-guided flood-fill algorithm. It is very expensive computationally. Zhang[15] proposed an enhanced two-frequency phase-shifting method. Similar to reference[14], the low-frequency fringe patterns are not with a unit frequency, so the phase of the low-frequency fringe patterns is wrapped. Zhang uses geometric constraints of DFP system to unwrap the low-frequency wrapped phase. The high-frequency wrapped phase is unwrapped with the low frequency continuous phase by the conventional dual-frequency temporal phase unwrapping. However, according to our own experience with it, Zhang's method has the following disadvantages: (1) this method not only needs to calibrate the digital fringe projection (DFP) system, but also needs to calculate the low frequency continuous phase corresponding to the nearest position of the measurement system, so the actual operation is relatively complex, (2) since geometric constraints can be used to unwrap the low-frequency wrapped phase, could it be directly use to unwrap the high-frequency one? If possible, the unwrapping process of the high-frequency phase can completed without projection, acquisition and phase demodulation of the low-frequency fringe pattern. The purpose of the present paper is to further improve Zhang's method. We present an improved phase-unwrapping method. This method is a good solution to overcome the issues of Zhang's method. The capability of the presented method is demonstrated by both theoretical analysis and experiments.

The remainder of this paper is as follows. Section 2 introduces the principle of the system. Section 3 presents the experimental results. Section 4 summarizes this paper.

## 2. Theory

### 2.1 Zhang's method

The conventional dual-frequency phase unwrapping algorithm use the continuous phase of the low-frequency fringe to unwrap the wrapped phase of the high-frequency one[15-16]. Assuming that the frequency of the low-frequency fringe is $f_1$, its phase is $\varphi_1(x,y)$, the frequency of the high-frequency fringe is $f_2$, its phase is $\varphi_2(x,y)$, $\varphi_1(x,y)$ is continuous phase when $f_1$ is less than 1. Thus $\varphi_2(x,y)$ can be unwrapped by the following equations,

$$k(x,y) = round[\frac{f_2\varphi_1(x,y)/f_1 - \varphi_2(x,y)}{2}]$$

$$\Phi_2(x,y) = \varphi_2(x,y) + 2k(x,y)\pi \qquad (1)$$

Where round[.] is the rounding operator, $\Phi_2(x, y)$ is the continuous phase corresponding to wrapped phase $\varphi_2(x, y)$. When the frequency ratio $f_2 / f_1$ between the high frequency and the low frequency is too large, there will be many errors in the final result. Zhang proposed an enhanced two-frequency phase-shifting method. Its main stages are briefly summarized as follows:

(1) Two sets of fringes with of the low and high frequency are projected on the tested object, then are captured by CCD camera. The period of the low-frequency fringe is larger than 1;

(2) The wrapped phases corresponding to the low and high frequency fringe are respectively obtained by phase demodulation algorithm;

(3) Calibrate the DFP system,calculate $\varphi_{\min}$ at a given depth $z = z_{\min}$ from geometric constraints of the system;

(4) The low-frequency wrapped phase is unwrapped using $\varphi_{\min}$, obtain the low frequency continuous phase ;

(5) The high-frequency wrapped phase is unwrapped with the low frequency continuous phase by Eq.(1),.

In Zhang's method,the noise impact can be substantially reduced by allowing the use of more than one period of low frequency phase map to determine the fringe order for the high frequency wrapped phase.

## 2.2 Our method

Though Zhang's method can reduce the noise impact of the dual-frequency phase unwrapping algorithm, it also has some disadvantages: (1) this method needs to calibrate the DFP system, calculate $\varphi_{\min}$ at a given depth $z = z_{\min}$ from geometric constraints of the system, so the actual operation is relatively complex, (2) since geometric constraints can be used to unwrap the low-frequency wrapped phase, could it be directly used to unwrap the high-frequency one? Based on our analysis of the work reported in Ref. [15], we propose an improved phase-unwrapping method using geometric constraints, and the fundamental idea is that the complex procedure of calibrating the DFP system and calculating $\varphi_{\min}$ at a given depth $z = z_{\min}$ is replaced by the simple experiment. Besides, only a set of high frequency fringe patterns is needed, the geometric constraints is used directly to unwrap the high-frequency wrapped phase.

In Ref. [15], $\varphi_{min}$ is calculated from geometric constraints of the DFP system, assuming that the wrapped phase is $\varphi_w$, as shown in Fig. 1.

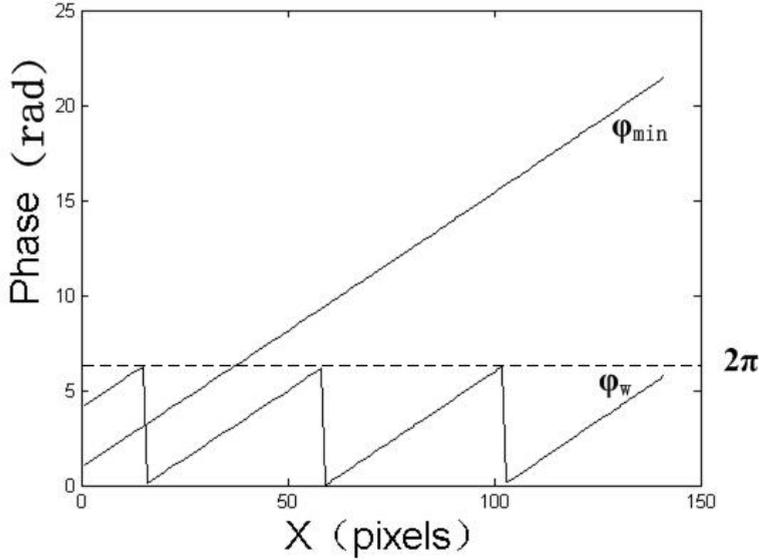

Fig.1 $\varphi_{min}$ and $\varphi_w$

If $0 < \varphi_{min} - \varphi_w < 2\pi$, $2\pi$ should be added; $2\pi < \varphi_{min} - \varphi_w < 4\pi$, $4\pi$ should be added; and $4\pi < \varphi_{min} - \varphi_w < 6\pi$, $6\pi$ should be added. That is,

$$K(x,y) = ceil[\frac{\varphi_{min} - \varphi_w}{2\pi}]$$
$$\Phi(x,y) = \varphi_w(x,y) + 2K(x,y)\pi \qquad (2)$$

Where ceil[] is the ceiling operator that returns the closest upper integer value. By our experiments, the high-frequency wrapped phase can be unwrapped by Eq.(2) after the low-frequency wrapped phase is unwrapped by $\varphi_{min}$, but the process of calibrating the DFP system and calculating $\varphi_{min}$ at a given depth $z = z_{min}$ is too complex, so we try to improve Zhang's method. On the basis of the original experiment, a reference plane is placed at $z = z_{min}$ (that is, the nearest position from the camera perspective) in DFP system, four-step shifting high frequency fringe patterns are respectively projected on the reference plane and tested object. The corresponding wrapped phases are obtained by four-step shifting algorithm. Because the reference plane is very smooth, its wrapped phase can be unwrapped easily and accurately by spatial phase unwrapping method[17-18], and the continuous phase on the reference plane is equivalent to $\varphi_{min}$ in [15]. Therefore, the $\varphi_{min}$ can be obtained by a simple measurement on the reference plane. Then the wrapped phase on the measured object would be unwrapped by Eq.(2). According to our experiments, most of the wrapped phase can be unwrapped, but sometimes there may be some wraps near the highest point when the frequency of the fringe pattern is too big or the tested object is too high, thus the partially unwrapped phase needs to be further corrected. Because the residual wraps is very little, the phase with residual wraps can be corrected easily. The specific procedure of correction is as follows. From the end of a line of phase data (such as the left), the corrected phase is taken as 0, the correction factor k=0; the phase values of two adjacent points are compared in the correction direction, if the phase difference is less than $-\pi$, $2\pi$ should be added to the corrected phase, and the correction factor k=k+1; if the phase difference is larger than $\pi$, $2\pi$ should be subtracted from the corrected phase, and the correction factor k=k-1, as shown in Eq.(3),

$$\Phi_K(i) = \Phi_U(i) + 2\pi K_i$$

$$K_i = \begin{cases} K_{i-1} & -\pi \leq \Phi_U(i) - \Phi_U(i-1) \leq \pi \\ K_{i-1} + 1 & \Phi_U(i) - \Phi_U(i-1) < -\pi \\ K_{i-1} - 1 & \Phi_U(i) - \Phi_U(i-1) > \pi \end{cases} \qquad (3)$$

$K_0 = 0$, $\Phi_u$ is the phase to be corrected, and $\Phi_K$ is the phase after correction.

The main stages of our algorithm are summarized as follows(as shown in Fig.2):
(1) project the four-step shifting high-frequency fringe onto the object's surface and capture the image by a CCD camera, as shown in Fig.2-A;
(2) project the four-step shifting high-frequency fringe onto the reference plane (which is placed at the nearest position from the camera perspective) and capture the image, as shown in Fig.2-B;
(3) obtain the wrapped phase corresponding to Fig.2-A, as shown in Fig.2-C;
(4) obtain the continuous phase of the reference plane by the spatial phase unwrapping algorithm, as show in Fig.2-D;
(5) unwrap the wrapped phase of measured object in Fig.2-C by Eq.(2),obtain the unwrapped phase of measured object, as shown in Fig.2-E, and to better visualize the details, we show the zoom-in view of local region,as shown in Fig.2-F. We can see that there still are some residual wraps after the phase unwrapping
(6) correct the phase with residual wraps in Fig.2-E by Eq.(3), obtain the continuous phase of the measured object as shown in Fig.2-G, and the zoom-in view is shown in Fig.2-H corresponding to Fig.2-F.
In step 5,if there still are some residual wraps in the unwrapped phase map of measured object, we do step 6, otherwise step 6 can be omitted.

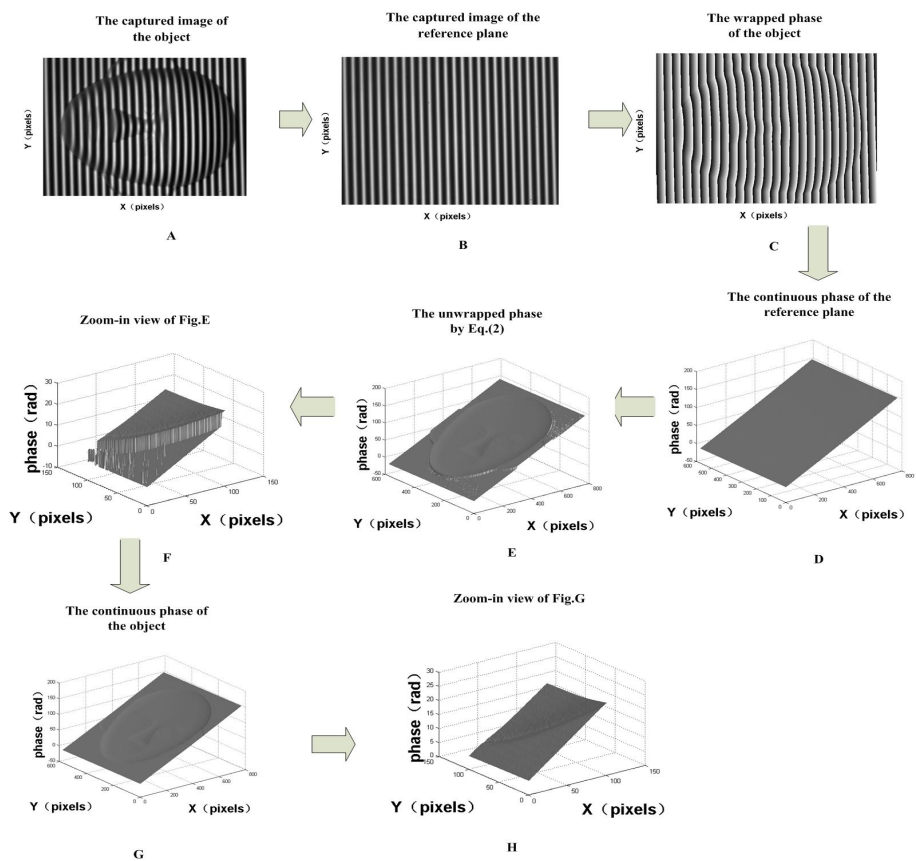

Fig. 2 The flowchart of processing pipeline

The following experiment is used to verify the proposed algorithm.

## 3. Experiments

In this section, for evaluating the real performance of our method, we test our method on a series of experiments. Below, we will describe these experiments and practical suggestions for the above procedure.

We develop a 3D shape measurement system, which consists of a DLP projector (Optoma EX762) driven by a computer and a CCD camera ( DH-SV401FM).The camera is attached with a 25mm focal length lens(Model: ComputarFA M2514-MP2). Fig.3 shows the optical path of phase measuring profilometry, where $P$ is the projection center of the projector, $C$ is the camera imaging center, and $D$ is an arbitrary point on the tested object. In our experiments, the distance between the camera and the projector is about 30 cm and the tested object is placed in front of the projector about 70 cm. The surface measurement software is programmed by Matlab with I5-4570 CPU @ 3.20 GHz.

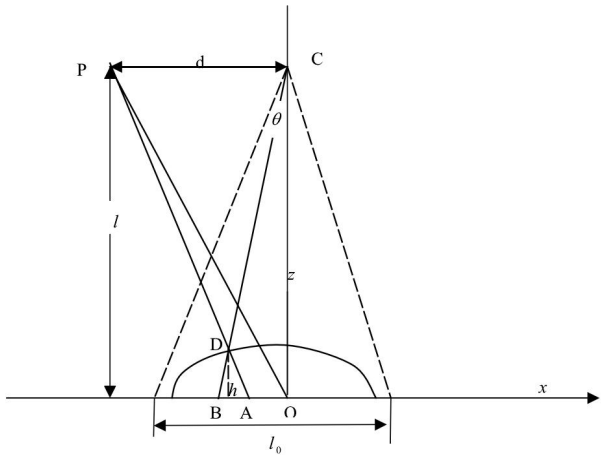

Fig. 3 Optical path of phase measuring profilometry

Firstly, an experiment is provided to demonstrate the feasibility of the proposed algorithm. The tested object is a face model and an isolated cup. Project the the four-step shifting high-frequency fringe onto the object's surface and reference plane respectively, the deformed fringe pattern is captured by a CCD camera as shown in Fig.4. Fig.4(a) is the captured image of the tested object and Fig.4(b) is the captured image of the reference plane. The captured image is 624 pixels wide by 441 pixels high.

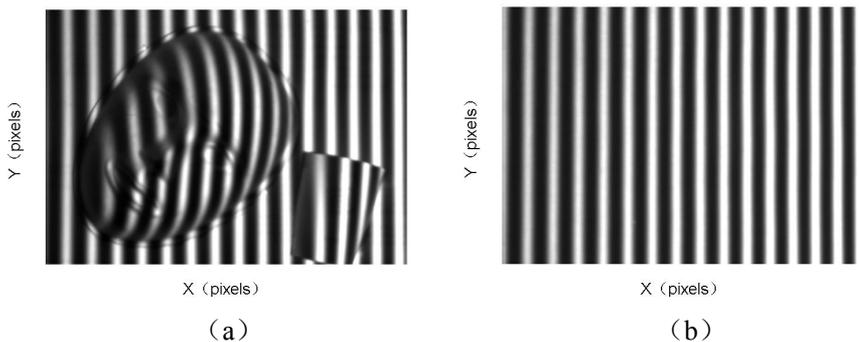

（a） （b）

Fig.4 The captured image

Then obtain the wrapped phase of the tested object and the reference plane using the four-step phase shifting algorithm, as shown in Fig.5. Fig.5(a) is the wrapped phase of the tested object, and Fig.5(b) is the wrapped phase of the reference plane.

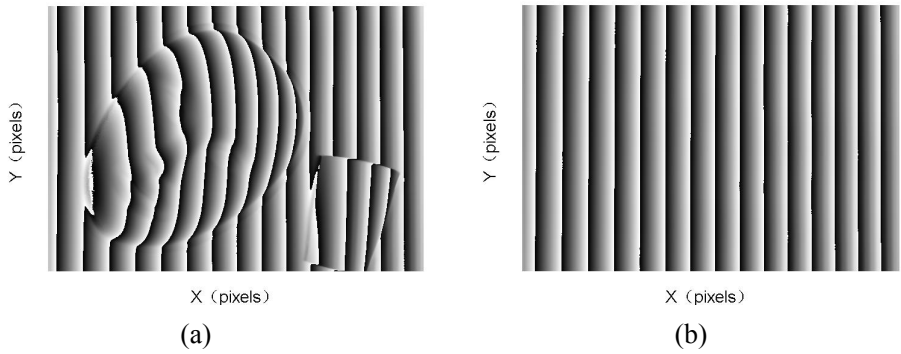

(a)  (b)

Fig.5 The wrapped phase

Unwrap the wrapped phase of the reference plane using the spatial phase unwrapping algorithm, Thus the minimum phase map $\varphi_{min}$ is obtained, which is shown in Fig.6.

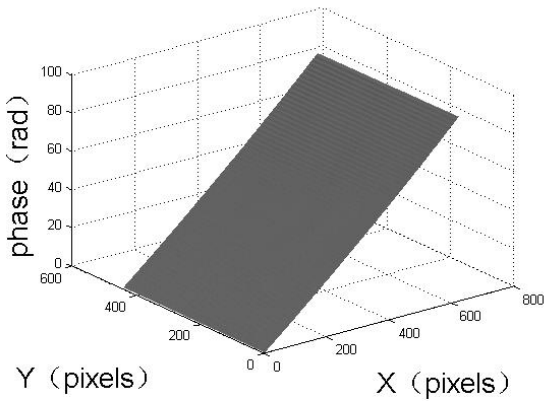

Fig.6 The continuous phase of the reference plane

Then the wrapped phase of the measured object(Fig.5(a)) is unwrapped by Eq.(2), the result is shown in Fig.7.

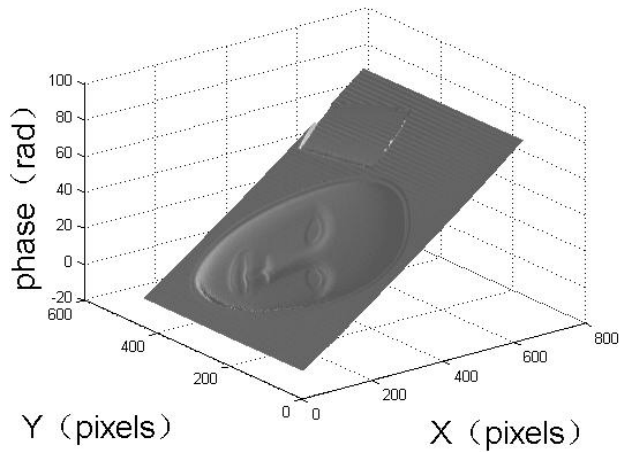

Fig.7 The continuous phase

For a more complete verification to the proposed method, we do another experiment on a plastic board with two big holes. The experiment procedure is similar to the first experiment.

Project the four-step shifting high-frequency fringe onto the object's surface and the reference plane respectively, the deformed fringe pattern is captured by a CCD camera as shown in Fig.8. Fig.8(a) is the captured image of the tested object and Fig.8(b) is the captured image of the reference plane. The captured image is 624 pixels wide by 441 pixels high.

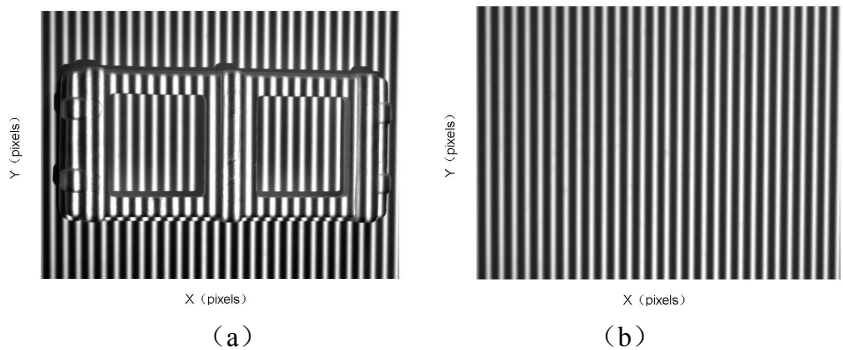

（a） （b）

Fig.8 The captured image

Then obtain the wrapped phase of the tested object and the reference plane estimated by he four-step phase shifting algorithm, as shown in Fig.9. Fig.9(a) is the wrapped phase of the tested object, and Fig.9(b) is the wrapped phase of the reference plane.

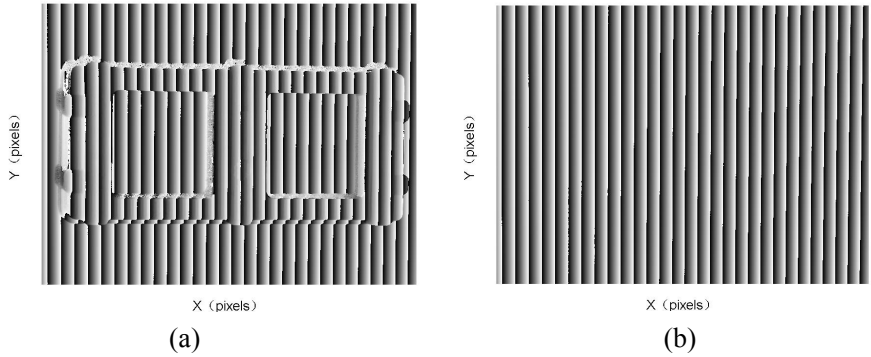

(a) (b)

Fig.9 The wrapped phase

Unwrap the wrapped phase of the reference plane using the spatial phase unwrapping algorithm, Thus the minimum phase map $\varphi_{\min}$ is obtained, which is shown in Fig.10.

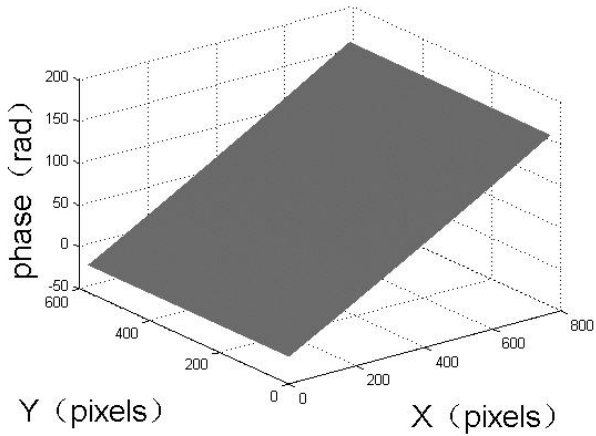

Fig.10 The continuous phase of the reference plane

Then the wrapped phase of the measured object(Fig.9(a)) is unwrapped by Eq.(2), the result is shown in Fig.11.

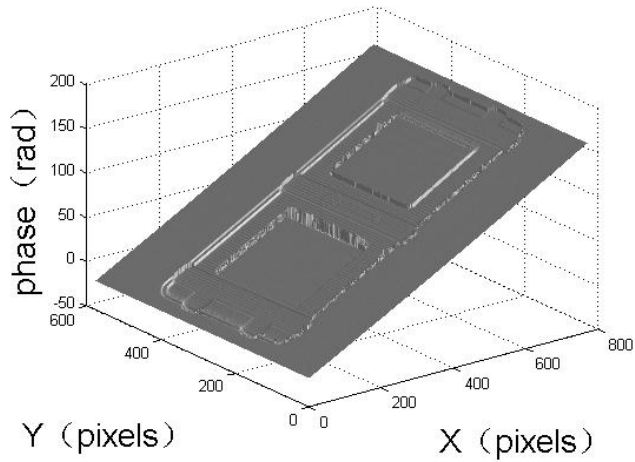

Fig.11 The continuous phase

Through the experiments, we found that the proposed method is more single and easier to realize than Zhang's method because it only needs a single high-frequency wrapped phase of tested object,,eliminates the complex calibration process of the DFP system,omits calculating of the minimum phase map $\varphi_{min}$. However, when the frequency of the fringe is too big or the tested object is too high, there may be some residual wraps near the highest point, which needs to be corrected. Because the residual wraps is very little, correction is easy.

## 4. Conclusion

In this paper, we propose an improved phase-unwrapping method using geometric constraints. This algorithm is basically an extension of Zhang's method.The proposed method eliminates the complex procedure of calibrating the DFP system and calculating the minimum phase map $\varphi_{min}$. The geometric constraints is used directly to unwrap the high-frequency wrapped phase. Only one single high-frequency fringe pattern is needed by the proposed method. The experiments results demonstrate that the proposed method can simplifies the process.

*Acknowledgment* - This work was supported by the National Natural Science Foundation of China(NSFC)( Grant Nos:11302082, 11472070 and 11672162). The support is gratefully acknowledged.